\documentclass[
aps,
prl,
twocolumn,
superscriptaddress,
amsmath,
amssymb,
floatfix,
longbibliography,
10pt
]{revtex4-1}

\usepackage{graphicx}
\usepackage{dcolumn}
\usepackage{bm}
\usepackage{stmaryrd}
\usepackage{latexsym}
\usepackage{amsfonts}
\usepackage{textcomp}

\begin{document}

\title{Out-of-distribution Neural Inference in Dynamical Ising Models}

\author{Yuan-Bin Zhu}
\affiliation{Center for Quantum Physics and Intelligent Sciences, Department of Physics, Capital Normal University, Beijing 10048, China}

\author{Shuang Qiao}
\email[Corresponding author. Email: ]{qiaos@cnu.edu.cn}
\affiliation{Center for Quantum Physics and Intelligent Sciences, Department of Physics, Capital Normal University, Beijing 10048, China}

\author{Shi-Ju Ran}
\email[Corresponding author. Email: ]{sjran@cnu.edu.cn}
\affiliation{Center for Quantum Physics and Intelligent Sciences, Department of Physics, Capital Normal University, Beijing 10048, China}

\date{\today}

\begin{abstract}
Neural networks are increasingly used to infer hidden physical structure from dynamical observations, yet it remains unclear whether their out-of-distribution performance reflects transferable physical rule learning. We address this question in a controlled inverse problem: reconstructing interaction graphs of a kinetic Ising model from Glauber magnetization trajectories. Across convolutional, graph, Transformer, and hybrid architectures, we find that data-driven training produces distinct and reproducible statistical strategies under topology and temperature shifts. Edge-population diagnostics reveal that Transformer-based models tend to preserve the link density of the training ensemble, whereas convolutional models can collapse toward sparse- or no-link predictions that appear out-of-distribution stable by exploiting the majority no-link class. Thus, high in-distribution accuracy and apparent out-of-distribution robustness do not necessarily imply a learned dynamics-to-structure rule. Instead, neural reconstruction can be governed by architecture-dependent statistical priors. Our results identify a concrete failure mode of standard data-driven learning in physical inverse problems and motivate rule-guided principles for machine-learning-assisted scientific discovery.
\end{abstract}

\maketitle

\paragraph*{Introduction.---} Artificial intelligence (AI) and machine learning (ML) have become increasingly important tools in, e.g., physics
\cite{A1,A2,A3}, biology \cite{B1,B2,B3}, materials science \cite{C1,C2}, and climate research
\cite{A3,D2,D3}. Beyond accelerating interpolation within existing data, a central ambition of AI for Science is to assist scientific discovery by predicting structures or mechanisms, in regimes that are not represented in the training data \cite{C1,F2,F3}. This ambition makes out-of-distribution (OOD) prediction a necessary benchmark for scientific ML: discovery requires extrapolation beyond the statistical ensemble from which the model has learned, not merely accurate prediction on new samples drawn from the same distribution  \cite{C1,H4,Add3,Add4,Add5}.

This requirement is in tension with the standard data-driven learning paradigm. Most supervised ML models are trained by minimizing empirical prediction error on finite datasets and are validated under an independently and identically distributed assumption. Under this setting, high in-distribution (ID) accuracy primarily demonstrates interpolation within the training ensemble. It does not, by itself, establish that the model has learned a transferable physical rule. In scientific applications, however, the relevant target often lies outside the training ensemble: a new material composition, a new phase, a new interaction topology, or a new dynamical regime. The key question is therefore not only whether neural networks predict accurately, but what kind of inference they perform when the test system violates the statistical ID assumptions of the training data.

The distinction between statistical interpolation and transferable physical inference has been widely discussed in the context of distribution shift. Studies in images, language, and benchmark learning tasks have shown that models with high ID performance can fail when the test distribution changes \cite{H1,H2,H3,H4,H5}. Similar concerns arise in physical applications, including materials property prediction \cite{H4}, quantum dynamics learning \cite{F2}, and complex-flow modeling \cite{L1}, where distribution shifts may be induced by changes in structure, control parameters, or regions of state space \cite{H4,L1,F2,K2}. For inverse problems, this issue is even sharper: the model must infer hidden physical structures or parameters from finite observations, and an accurate prediction may result either from a learned physical relation or from a statistical regularity inherited from the training ensemble \cite{N1,N2,O1,O2}.

\begin{figure}[t]
\centering
\includegraphics[width=0.48\textwidth]{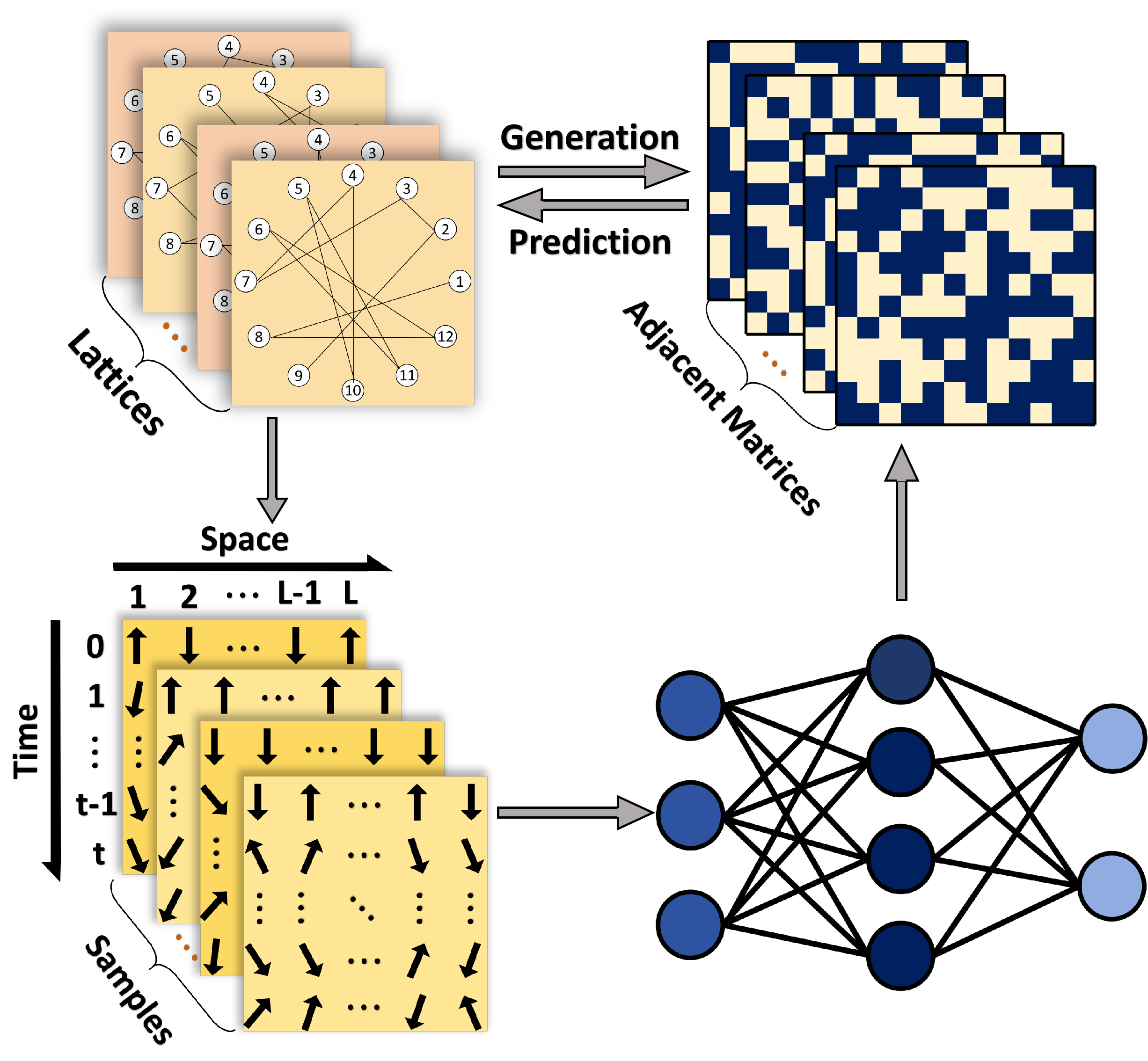}
\caption{
Schematic illustration of topology reconstruction in the kinetic Ising model using neural-network models.
Given a time-dependent local magnetization trajectory, the model predicts the underlying adjacency matrix of the
interaction graph.
}
\label{fig:fig1}
\end{figure}

Here we address this issue using the kinetic Ising model with Glauber dynamics as a controlled physical testbed \cite{Add1}. The forward problem maps an interaction topology, represented by an adjacency matrix, to time-dependent local magnetization trajectories. The inverse problem considered here is to reconstruct the underlying interaction topology from these trajectories (Fig.~\ref{fig:fig1}). This allows us to separate three key notions that are often conflated in scientific ML: ID fitting, OOD prediction, and transferable dynamics-to-structure inference. The training and ID test data are generated from the same ensemble of lattice topologies and temperatures. We consider OOD generalization under two physically distinct distribution shifts: the topology shift with the lattice structures absent from training, and the temperature shift where the trajectories are generated at unseen temperatures. We compare convolutional neural networks (CNNs), graph neural networks (GNNs), Transformers, and hybrid architectures, which encode different inductive biases for processing spatiotemporal magnetization data.

Our results reveal that data-driven neural reconstruction in this physical inverse problem is governed by architecture-dependent inference strategies under distribution shift. Although different architectures usually achieve high in-distribution accuracy, which is a widely-recognized fact, they respond differently when the graph topology or dynamical temperature is changed. Edge-population diagnostics show that Transformer-based models tend to preserve the link density of the training ensemble, whereas convolutional models can collapse toward sparse-link or no-link predictions that exploit the majority no-link class. These behaviors show that different neural architectures do not merely differ in accuracy; they implement distinct statistical strategies for mapping dynamics to structure. This positive diagnostic finding leads to an important caution: apparent OOD robustness need not imply a transferable physical rule, but may instead arise from architecture-dependent statistical priors. The results identify a concrete failure mode of standard supervised learning in physical inverse problems and motivate mechanism-aware, rule-guided approaches for machine-learning-assisted scientific discovery.

\paragraph*{Kinetic Ising model as a testbed for generalization.---}
We consider a high-dimensional Ising model with interaction topology encoded by an adjacency matrix \(\bm{A}\),
\begin{equation}
H=-J\sum_{i,j} A_{ij}s_i s_j.
\end{equation}
The local magnetization dynamics governed by the mean-field Glauber equation
\begin{equation}
\frac{dm_i}{dt}=-m_i+\tanh\Big(\beta J\sum_j A_{ij}m_j\Big),
\end{equation}
where \(m_i(t)=\langle s_i(t)\rangle\) (note we take the Boltzmann constant as \(k_B=1\)). For each topology, we numerically solve Eq.~(2) from sampled initial conditions and use the resulting magnetization trajectories as neural-network inputs. The prediction target is the upper-triangular part of \(\bm{A}\), corresponding to all candidate undirected links.

We evaluate the ID and two OOD test sets. The ID set uses the same topology ensemble and temperature as training, with independently sampled initial conditions. The topology-shift OOD set uses unseen lattice structures at fixed temperature, whereas the temperature-shift OOD set uses training topologies but generates trajectories at unseen temperatures. These settings separately probe structural and thermodynamic distribution shifts.

We compare two-layer CNN (CNN-2), a deeper three-layer CNN (CNN-3), GNN, Transformer, and hybrid CNN-Transformer architectures, which encode convolutional locality \cite{P2,Q2}, message-passing relational structure \cite{R1}, attention-based long-range dependence \cite{P2,S2}, and hybrid local--global processing \cite{P2,T2}, respectively. The details for different architectures are given in the Supplemental Material~\cite{Supp}. All models are trained for \(L=12\) spins, giving \(L(L-1)/2=66\) candidate links. The main setting uses \(N_c=25\) true links, for which a no-link predictor already reaches \(41/66=62.1\%\). We therefore report both reconstruction accuracy \(\gamma\) and the average predicted link number \(\widehat{N}_{c}\). A balanced \(N_c=33\) with temperature shift is explored in the Supplemental Material~\cite{Supp}. Training uses Adam optimizer~\cite{Add2} and binary cross-entropy loss,
\begin{equation}
\mathcal{L}=-\frac{1}{N}\sum_{i=1}^{N}
\left[y_i\log(\widehat{y}_i)+(1-y_i)\log(1-\widehat{y}_i)\right].
\end{equation}

\begin{figure*}[t]
\centering
\includegraphics[width=0.88\textwidth]{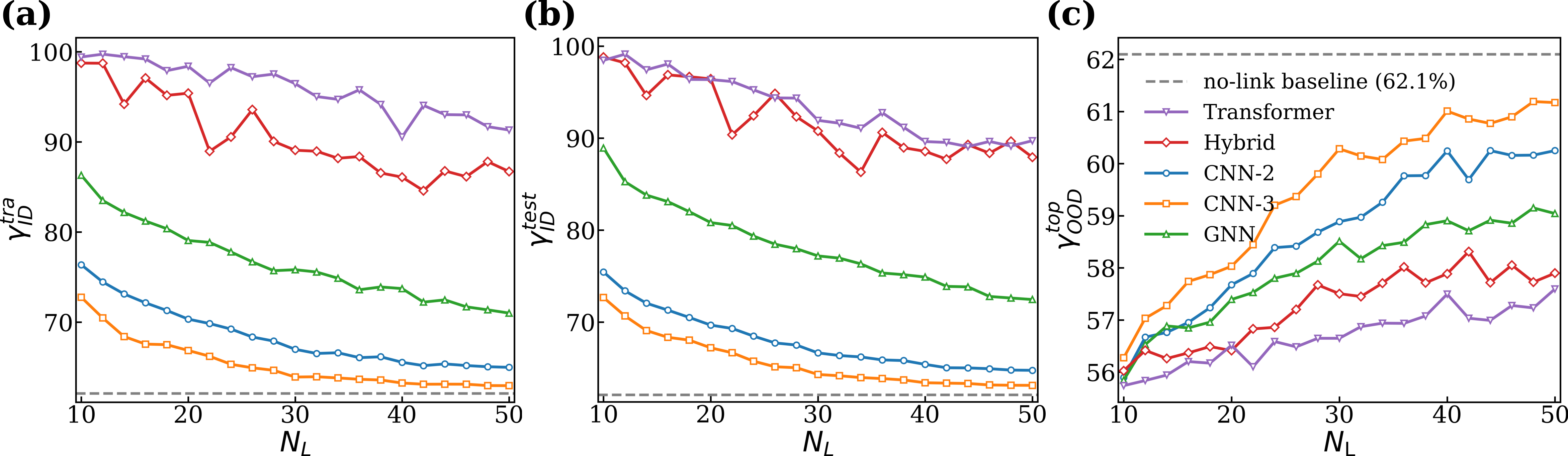}
\caption{
(a) Training accuracy \(\gamma_{\mathrm{ID}}^{\mathrm{tra}}\),
(b) ID test accuracy \(\gamma_{\mathrm{ID}}^{\mathrm{test}}\), and
(c) topology-shift OOD accuracy \(\gamma_{\mathrm{OOD}}^{\mathrm{top}}\) of the five models as a function of the
number of training lattice topologies \(N_L\). The dashed line labels the accuracy of the no-link baseline, \(41/66=62.1\%\).
}
\label{fig:fig2}
\end{figure*}

\paragraph*{OOD generalization under topology shift.---} We first evaluate under a topology shift. The training set contains \(N=35000\) trajectories generated at \(T=1\)  from \(N_L\) lattice topologies, with \(N/N_L\) trajectories sampled from each topology. The ID test set uses the same topology ensemble and temperature but independently sampled initial conditions, whereas the OOD set replaces the training topologies by unseen lattices at the same temperature.

Figures~\ref{fig:fig2}(a) and \ref{fig:fig2}(b) show that all five neural networks reach high training and ID test accuracies, with similar trends as \(N_L\) varies. Generally, the ranking is Transformer, Hybrid, GNN, CNN-2, and CNN-3, indicating that Transformer-based models extract the most predictive information within the ID setting. The decrease of ID accuracies with increasing \(N_L\) is consistent with the fixed sample budget: increasing \(N_L\) both reduces the samples per topology and broadens the set of topology-dependent dynamical responses, thereby making the inverse problem harder.

The same models behave qualitatively differently under topology shift. As shown in Fig.~\ref{fig:fig2}(c), all accuracies drop substantially; for \(N_L=10\), they are only around \(56\%\). The ranking is also reversed: CNN-2 and CNN-3 outperform the Transformer. Moreover, the OOD accuracy increases with \(N_L\), with CNN-3 rising from about \(56\%\) at \(N_L=10\) to about \(61\%\) at \(N_L=50\), opposite to the ID trend.

These results show that ID and topology-OOD tests probe different aspects of the learned inverse map. ID accuracy primarily measures trajectory-level interpolation: the model is evaluated on unseen dynamical realizations generated from the same topology ensemble used for training. In this regime, increasing \(N_L\) makes the reconstruction task harder, and the accuracy decreases toward, but remains above, the no-link baseline.

By contrast, the topology-OOD test probes topology-level extrapolation, requiring the same learned map to reconstruct interaction graphs that are absent from the training ensemble. The opposite dependence on \(N_L\) therefore indicates that the inductive biases favoring ID interpolation do not necessarily favor transferable dynamics-to-topology inference. More importantly, it shows that OOD accuracy alone is not a direct measure of physical rule learning: it must be interpreted together with reconstruction diagnostics that reveal what type of graph the model actually predicts. This motivates a complementary diagnostic beyond edge-wise accuracy.

\paragraph*{OOD generalization under temperature shift.---} We next test temperature shift by training all models at \(T=1\) with \(N=35000\) trajectories and \(N_L=50\) topologies, and evaluating them on the same topologies at unseen temperatures  ranging from \(T=0.05\) to \(100\). This setting keeps the interaction graphs fixed while changing the dynamical trajectories generated by the Glauber equation.

\begin{figure*}[t]
\centering
\includegraphics[width=0.88\textwidth]{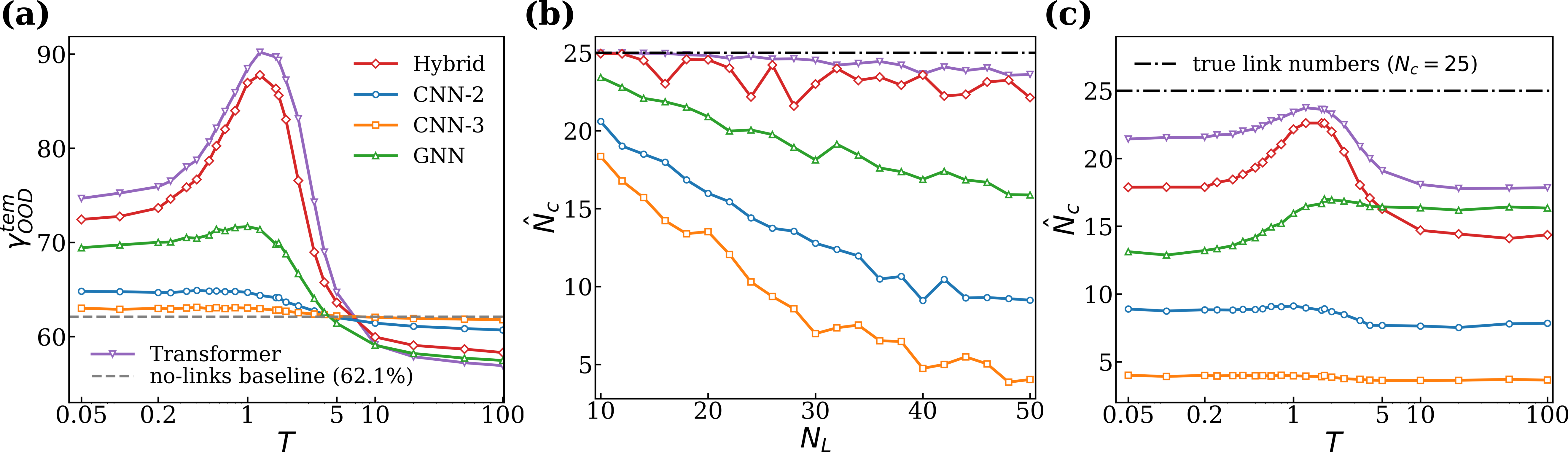}
\caption{
(a) Temperature-shift OOD accuracy \(\gamma_{\mathrm{OOD}}^{\mathrm{tem}}\) of the five models as a function of
\(T\). The dashed line labels the accuracy of the no-link baseline, \(41/66=62.1\%\).
(b) Average number of predicted links \(\widehat{N}_c\) in the reconstructed adjacency matrices on the ID test
set as a function of \(N_L\).
(c) Average number of predicted links \(\widehat{N}_c\) on the temperature-shift OOD test set as a function of
\(T\) in the \(N_c=25\) setting. The dash-dotted lines indicate the true link number \(N_c=25\).
}
\label{fig:fig3}
\end{figure*}

Figure~\ref{fig:fig3}(a) shows an asymmetric response to low- and high-temperature shifts. Near the training temperature \(T=1\), all models retain high accuracy and preserve the ID ranking. In the low-temperature regime, \(T=0.05\)--\(0.2\), the accuracy decreases moderately while the ranking remains unchanged, with the Transformer dropping from about \(90\%\) at \(T=1\) to about \(75\%\) at \(T=0.05\), whereas CNN-3 remains near \(63\%\). The high-temperature regime is more disruptive. As \(T\) increases, all models degrade, and a crossover occurs around \(T\simeq 5\)--\(10\), after which the ranking becomes CNN-3, CNN-2, Hybrid, GNN, and Transformer. In this regime, the accuracies fall below the no-link baseline, indicating that temperature shift can destroy link-specific predictive information rather than merely perturb the input distribution.


The stronger degradation under high-temperature shifts than under low-temperature shifts has a direct physical origin. Lowering \(T\) enhances magnetic ordering and shifts the trajectory distribution, but topology-dependent correlations remain partly visible in the ordered dynamics. Raising \(T\), by contrast, suppresses magnetic ordering and weakens topology-dependent dynamical responses, thereby reducing the information available for link reconstruction. Temperature shift therefore probes more than robustness to a changed input distribution: it tests whether the learned inverse map can still extract topology-relevant signals when the dynamical signatures of the interaction topology are weakened by thermal fluctuations. 

\paragraph*{Prediction strategies behind apparent OOD robustness.---}
The preceding OOD tests reveal a central ambiguity: similar accuracy values can arise from qualitatively different reconstruction strategies. A model may achieve high accuracy by identifying the correct links, by preserving the typical link density of the training ensemble, or by exploiting the majority no-link class. This ambiguity is especially important in the \(N_c=25\) setting, where 41 of the 66 candidate pairs are unlinked and a naive no-link predictor already reaches \(41/66=62.1\%\). Thus, the nearly flat OOD accuracy of CNN-3 and the strong degradation of the Transformer cannot be interpreted from accuracy alone. This calls for a diagnostic of what type of graph each model actually predicts.

We use the average predicted number of links, \(\widehat{N}_c\), as such a diagnostic (Figs.~\ref{fig:fig3}(b) and \ref{fig:fig3}(c)). This quantity separates models that preserve the training graph density from those that collapse toward sparse-link or no-link predictions. On the ID test set, the Transformer largely preserves the graph density, with \(\widehat{N}_c\) decreasing only from about 25 at \(N_L=10\) to about 23 at \(N_L=50\). CNN-3 behaves differently: \(\widehat{N}_c\) drops from about 18 to about 3 over the same range. This contrast persists under temperature shift at \(N_L=50\). Over \(T=0.05\)--\(100\), CNN-3 remains near \(\widehat{N}_c\simeq 3\), whereas the Transformer continues to predict many links, with \(\widehat{N}_c\simeq 21\) at \(T=0.05\), \(\widehat{N}_c\simeq 23\) at \(T=1\), and \(\widehat{N}_c\simeq 18\) for \(T\gtrsim 10\).

These diagnostics reveal two distinct prediction strategies. The Transformer follows a density-preserving strategy: it predicts a graph density close to that of the training ensemble, which supports high ID accuracy when the learned correlations also localize links correctly. Under OOD shifts, however, this strategy becomes fragile because many predicted links are placed at incorrect positions. CNN-3 instead moves toward a conservative no-link majority strategy in the imbalanced task. Its relatively flat OOD accuracy mainly reflects correct classification of many no-link entries, not reliable recovery of true links.

This interpretation also consistently clarifies the topology-shift results in Fig.~\ref{fig:fig2}(c). Increasing \(N_L\) can reduce overfitting to a small set of graph realizations, but it also makes some models more conservative, as indicated by the decreasing \(\widehat{N}_c\). In the imbalanced \(N_c=25\) setting, such conservativeness can raise apparent accuracy by suppressing false positives on the majority no-link class. Since the OOD accuracy remains close to or below the no-link baseline, however, this improvement should not be interpreted as strong knowledge transfer. A balanced-density control with \(N_c=33\), reported in the Supplemental Material~\cite{Supp}, confirms that the low-link behavior of CNN-3 is shaped by the interplay between architectural bias and the class prior.

These results sharpen the interpretation of OOD robustness in physical inverse problems. In the kinetic Ising reconstruction task, stable OOD accuracy can arise either from transferable dynamics-to-structure inference or from statistical strategies such as preserving the training graph density and exploiting the majority no-link class. The predicted link population, \(\widehat{N}_c\), exposes this distinction and shows that different architectures implement different inference strategies under the same physical shift. Thus, OOD accuracy alone is insufficient evidence for physical rule learning, and it must be accompanied by diagnostics of the predicted graph structure, consistent with broader observations that deep networks can exploit shortcut rules and that standard accuracy can obscure qualitatively different decision strategies \cite{U1,U2,U3}.

\paragraph*{Conclusion.---}
We have studied out-of-distribution neural reconstruction in a controlled physical inverse problem: inferring kinetic Ising interaction graphs from Glauber magnetization trajectories. Although all architectures achieve high in-distribution accuracy, topology and temperature shifts reveal that their OOD behavior is governed by distinct data-driven inference strategies. Edge-population diagnostics show that Transformer-based models tend to preserve the link density of the training ensemble, whereas convolutional models can collapse toward sparse-link or no-link predictions that exploit class imbalance. Thus, apparent OOD robustness does not necessarily imply transferable dynamics-to-structure rule learning. These results point to a broader open challenge for AI-assisted physics: how to combine data, architectures, and training objectives so that neural models move beyond dataset-specific statistical shortcuts toward rule-guided, knowledge-driven physical prediction.

\begin{acknowledgments}
This work was supported by the National Natural Science Foundation of China (Grant No.~12404092) and Beijing Natural Science Foundation (Grant No. QY25387). The numerical simulations were partially performed on the robotic AI-Scientist platform of Chinese Academy of Sciences.
\end{acknowledgments}

\bibliographystyle{apsrev4-2}
\bibliography{references}

\end{document}


\title{Supplemental Material for\\
``Out-of-distribution Neural Inference in Dynamical Ising Models''}

\author{Yuan-Bin Zhu}
\affiliation{Center for Quantum Physics and Intelligent Sciences, Department of Physics, Capital Normal University, Beijing 10048, China}

\author{Shuang Qiao}
\email[Corresponding author. Email: ]{qiaos@cnu.edu.cn}
\affiliation{Center for Quantum Physics and Intelligent Sciences, Department of Physics, Capital Normal University, Beijing 10048, China}

\author{Shi-Ju Ran}
\email[Corresponding author. Email: ]{sjran@cnu.edu.cn}
\affiliation{Center for Quantum Physics and Intelligent Sciences, Department of Physics, Capital Normal University, Beijing 10048, China}

\date{\today}

\nopagebreak
\maketitle

This file contains the following contents supplementary to the main text.

\begin{enumerate}
    \item Neural-network architectures.
    \item Balanced-link control with \(N_c=33\).
\end{enumerate}

\section{Neural-network architectures}
\label{app:architectures}

\begin{figure*}[t]
\centering
\includegraphics[width=0.95\textwidth]{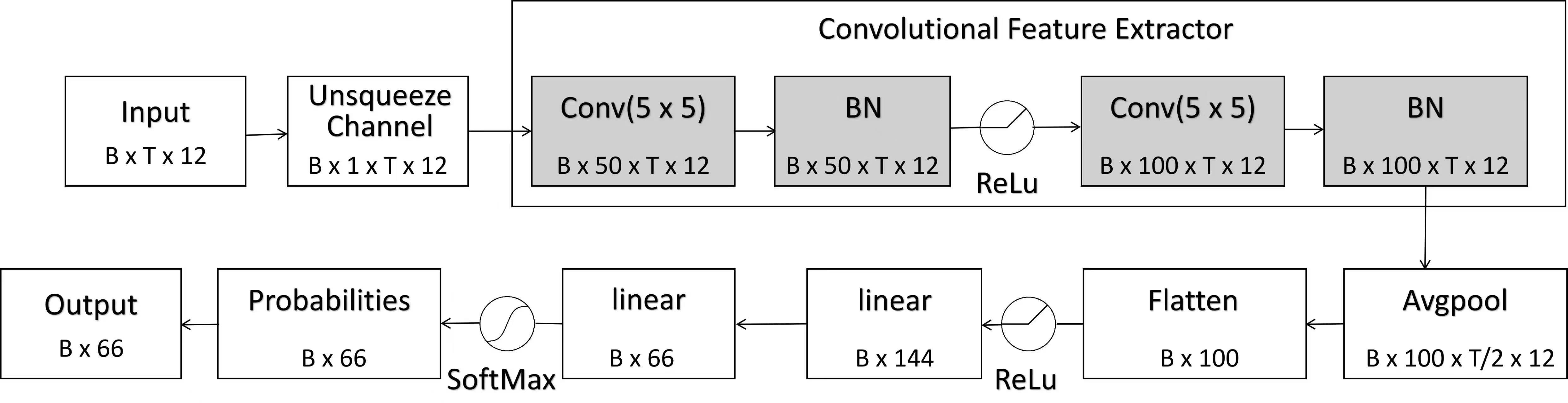}
\vspace{-6pt}
\caption{
Convolutional neural-network architecture used for topology reconstruction.
The input magnetization trajectory is processed by convolutional and batch-normalization layers,
followed by nonlinear activations and fully connected output layers that predict the upper-triangular
entries of the adjacency matrix.
}
\label{fig:app_cnn}
\end{figure*}

\begin{figure*}[t]
\centering
\includegraphics[width=0.95\textwidth]{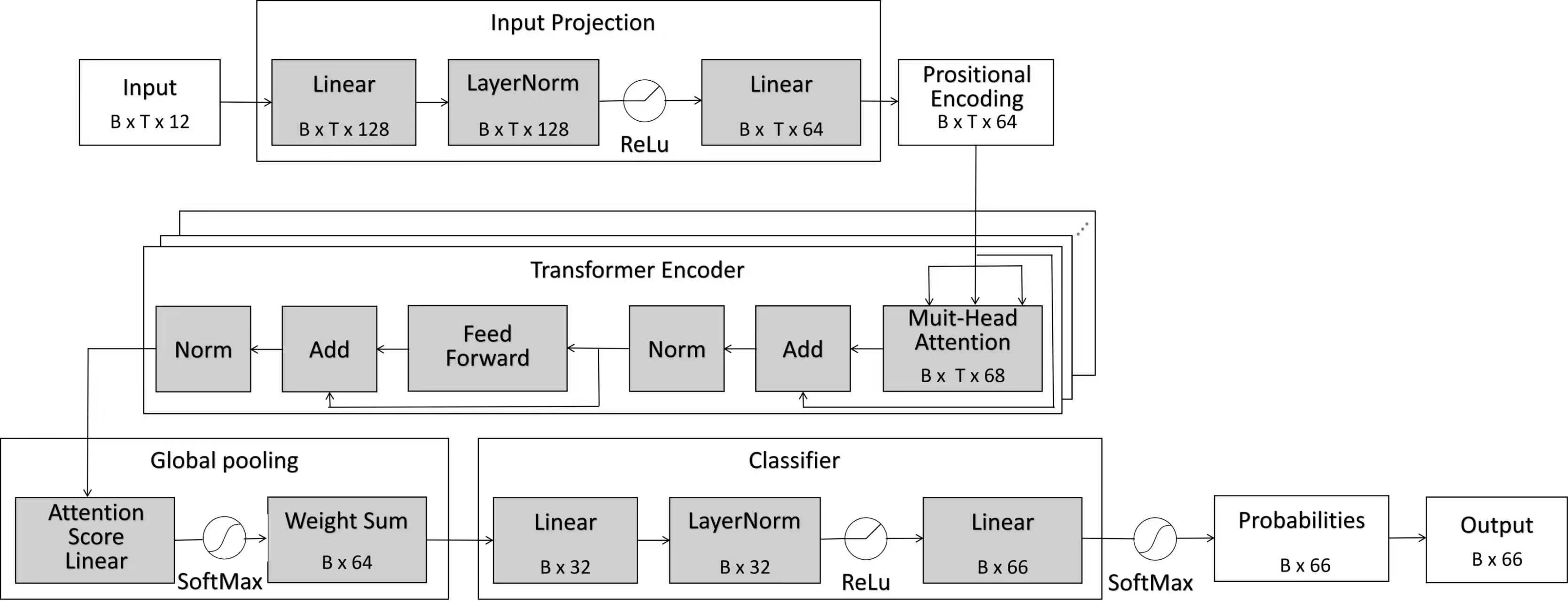}
\vspace{-6pt}
\caption{
Transformer-based architecture used for topology reconstruction.
The trajectory features are embedded by linear layers and processed by Transformer encoder blocks,
including multi-head attention, residual connections, normalization layers, and feed-forward layers.
The final linear layer outputs the predicted link probabilities.
}
\label{fig:app_transformer}
\end{figure*}

\begin{figure*}[t]
\centering
\includegraphics[width=0.95\textwidth]{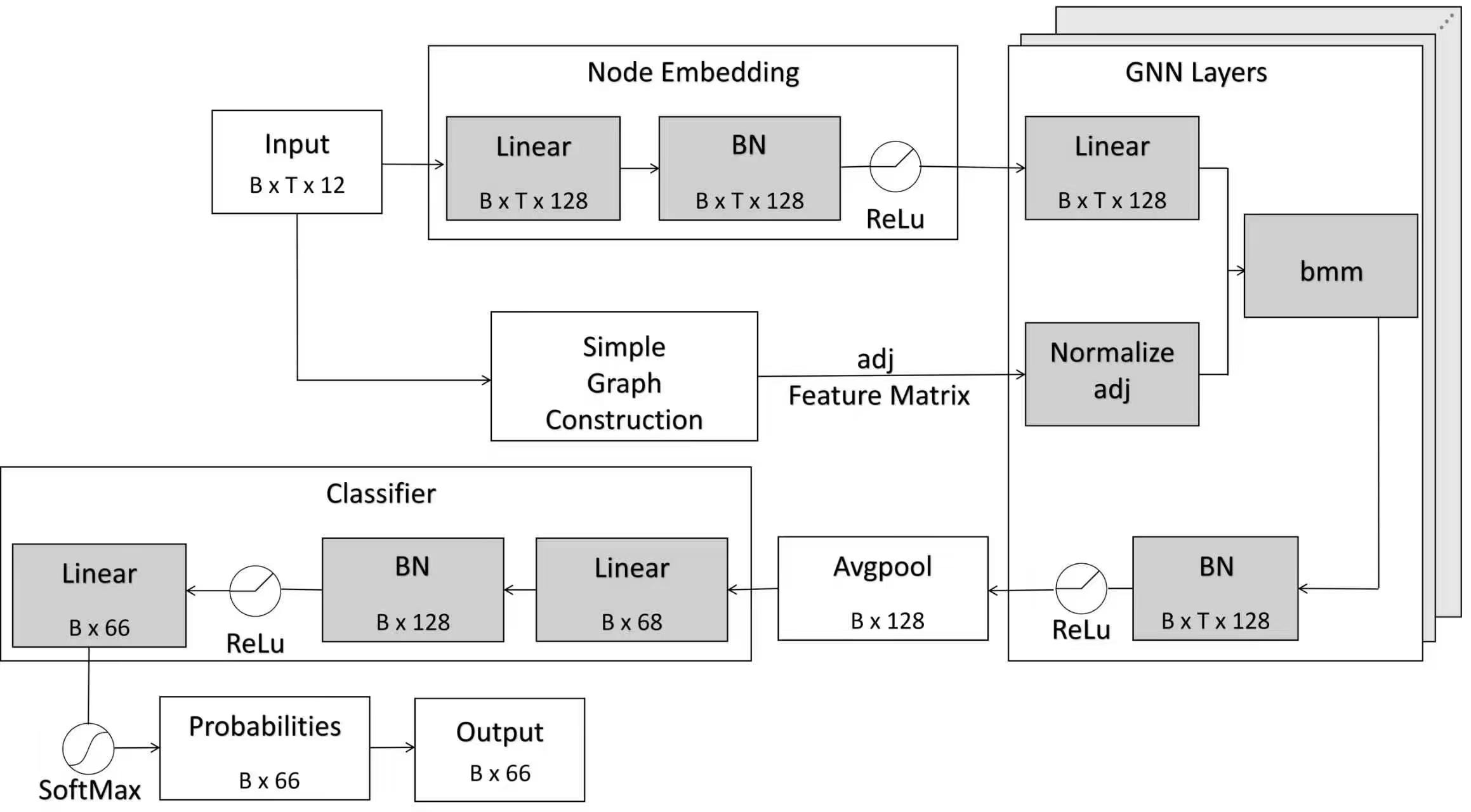}
\vspace{-6pt}
\caption{
Graph neural-network architecture used for topology reconstruction.
The model combines node-wise feature embedding, graph neural-network layers, normalization,
and linear readout layers to predict the interaction topology from magnetization trajectories.
}
\label{fig:app_gnn}
\end{figure*}

Figure~\ref{fig:app_cnn} shows the convolutional neural-network architecture used in this work.
Figure~\ref{fig:app_transformer} shows the Transformer-based architecture.
Figure~\ref{fig:app_gnn} shows the graph neural-network architecture.
These architectures are used to compare different inductive biases in topology reconstruction from Glauber magnetization trajectories.

\section{Balanced-link control with \(N_c=33\)}
\label{app:balanced}

\begin{figure*}[t]
\centering
\includegraphics[width=0.95\textwidth]{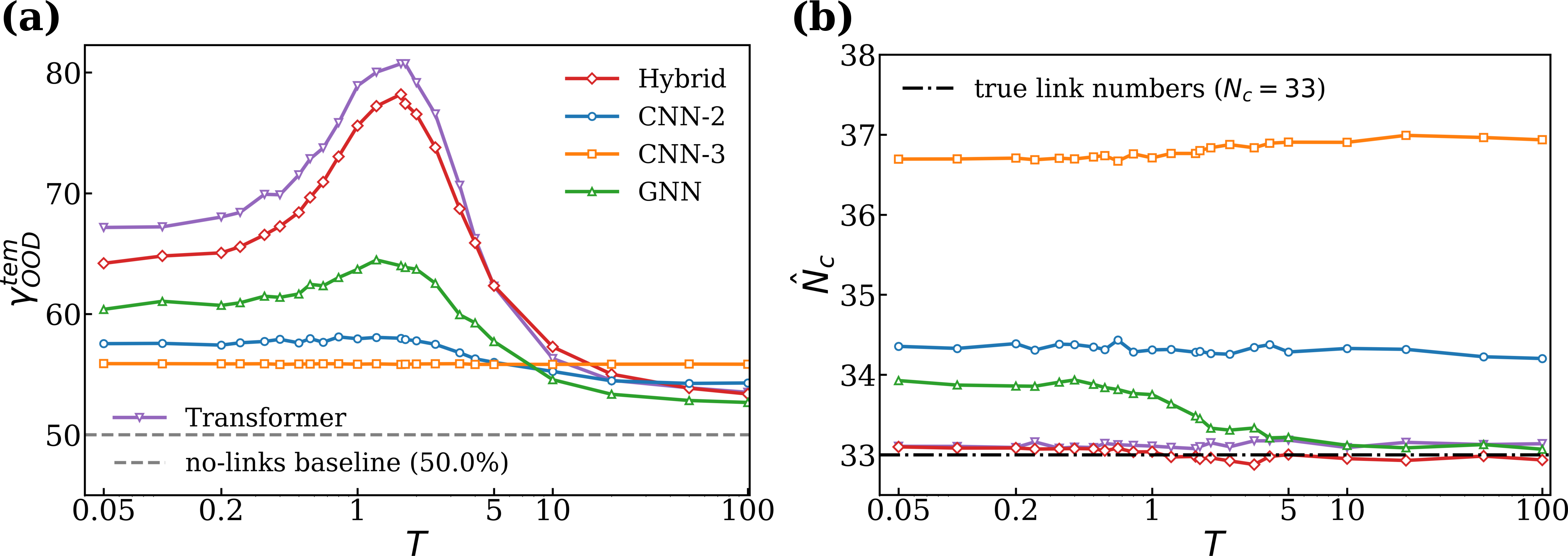}
\vspace{-6pt}
\caption{
Balanced-link temperature-shift control with \(N_c=33\).
(a) Temperature-shift OOD accuracy \(\gamma_{\mathrm{OOD}}^{\mathrm{tem}}\) of the five neural-network models as a function of \(T\).
The dashed line indicates the no-link baseline, which equals \(50.0\%\) in the balanced \(N_c=33\) setting.
(b) Average predicted number of links \(\widehat{N}_c\) as a function of \(T\).
The dash-dotted line indicates the true link number \(N_c=33\).
}
\label{fig:fig4}
\end{figure*}

Figure~\ref{fig:fig4} shows the temperature-shift OOD results for the balanced-link setting with \(N_c=33\).
Since \(L=12\) gives 66 candidate links, \(N_c=33\) removes the majority-class advantage of the no-link baseline, whose accuracy becomes \(50.0\%\).
Compared with the imbalanced \(N_c=25\) setting in the main text, this control demonstrates that the sparse-link behavior of CNN-3 is not solely an impact of architectural bias, but is affected by the relation between architectural bias and class prior.
In the balanced case, the predicted link number \(\widehat{N}_c\) remains close to the true density for several models over a broad temperature range, while the temperature-dependent accuracy still shows architecture-dependent degradation.
This result supports the conclusion that OOD accuracy should be interpreted together with reconstruction-density diagnostics.

